\documentclass[11pt]{article}

\usepackage[preprint]{acl}

\usepackage{times}
\usepackage{latexsym}

\usepackage[T1]{fontenc}

\usepackage[utf8]{inputenc}

\usepackage{microtype}

\usepackage{inconsolata}

\usepackage{graphicx}

\usepackage{listings, multicol}
\usepackage{color}
    \definecolor{codegray}{rgb}{0.5,0.5,0.5}
    \definecolor{backcolour}{rgb}{0.95,0.95,0.92}
    \lstdefinestyle{mystyle}{
        backgroundcolor=\color{backcolour},   
        numberstyle=\tiny\color{codegray},
        basicstyle=\ttfamily\tiny,
        breakatwhitespace=true,         
        breaklines=true,                 
        captionpos=b,                    
        keepspaces=true,                 
        numbers=none,               
        showspaces=false,                
        showstringspaces=false,
        showtabs=false,                  
        tabsize=2
    }
\usepackage{booktabs}
\usepackage{multirow}
\usepackage{amsmath}
\usepackage{amssymb}
\usepackage{enumitem}
    \setlist{noitemsep}
\usepackage{xcolor}
\usepackage{siunitx}
\usepackage[symbol]{footmisc}
    \newsavebox{\mybox}
\usepackage{makecell}

\title{AnalyticsGPT: An LLM Workflow for Scientometric Question Answering}

\author{Khang Ly, Georgios Cheirmpos, Adrian Raudaschl, 
\\ \textbf{Christopher James, Seyed Amin Tabatabaei} \\
        Elsevier B.V. \\ \{k.ly, g.cheirmpos, a.raudaschl, cd.james, s.tabatabaei\}@elsevier.com}

\begin{document}
\maketitle
\begin{abstract}
This paper introduces \textsc{AnalyticsGPT}, an intuitive and efficient large language model (LLM)-powered workflow for scientometric question answering. This underrepresented downstream task addresses the subcategory of meta-scientific questions concerning the ``science of science.'' When compared to traditional scientific question answering based on papers, the task poses unique challenges in the planning phase. Namely, the need for named-entity recognition of academic entities within questions and multi-faceted data retrieval involving scientometric indices, e.g. impact factors. 
Beyond their exceptional capacity for treating traditional natural language processing tasks, LLMs have shown great potential in more complex applications, such as task decomposition and planning and reasoning. 
In this paper, we explore the application of LLMs to scientometric question answering, and describe an end-to-end system implementing a sequential workflow with retrieval-augmented generation and agentic concepts. We also address the secondary task of effectively synthesizing the data into presentable and well-structured high-level analyses. 
As a database for retrieval-augmented generation, we leverage a proprietary research performance assessment platform. 
For evaluation, we consult experienced subject matter experts and leverage LLMs-as-judges. In doing so, we provide valuable insights on the efficacy of LLMs towards a niche downstream task. Our (skeleton) code and prompts are available at: \href{https://github.com/lyvykhang/llm-agents-scientometric-qa/tree/acl}{https://github.com/lyvykhang/llm-agents-scientometric-qa/tree/acl}.
\end{abstract}

\section{Introduction}\label{sec:introduction}
\begin{figure*}[]
    \centering
    \includegraphics[width=\linewidth]{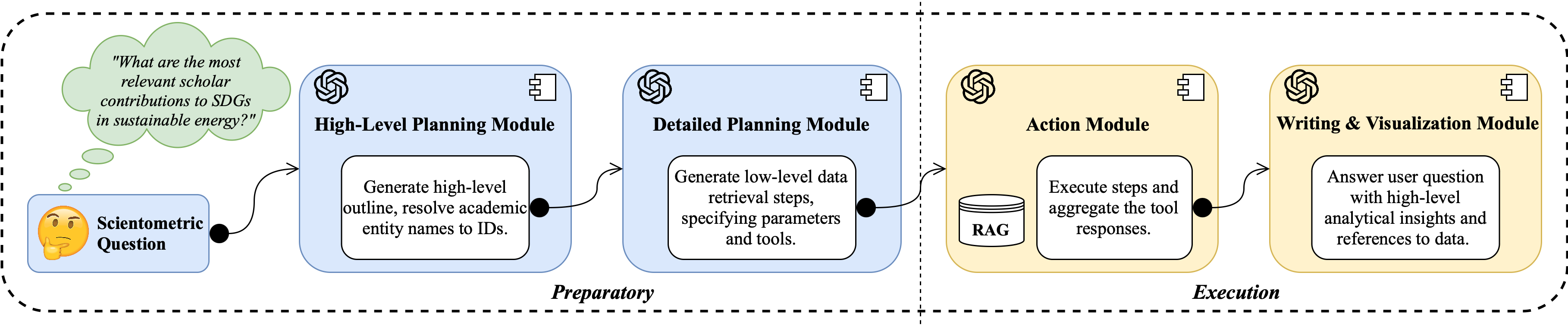}
    \caption{\label{fig:sys_diag}Overview of \textsc{AnalyticsGPT}, showing the main modules: High-Level Planning Module (HLPM), Detailed Planning Module (DPM), Action Module (AM), Writing Module (WM), and Visualization Module (VM). Each module, including user input semantics and the RAG interface, is further discussed separately in Section~\ref{sec:methodology}.}
\end{figure*}
Scientometric question answering (SQA) is essential to personal and institutional performance assessment, and can have far-reaching effects, e.g. the reputation and credibility of an author, or the attractiveness of an institution to a prospective student. For instance, annual university rankings based on scientometric data, e.g. by Quacquarelli Symonds or Times Higher Education, are highly influential within the higher education sector; such rankings primarily measure research productivity, and are employed for comparative analysis by a variety of stakeholder groups - potential customers, researchers, government agencies and policymakers, to name several~\citep{altbach2012, hazelkorn2015}. 
Citation databases are centralized sources of literature and impact metrics, and serve as an authoritative, objective criterion of scientific activities~\citep{gevorgyan2016}; such databases can serve as rich knowledge bases for SQA tasks.

\par Given their vast internal knowledge and approximation of human pattern recognition, the deployment of LLM-controlled agents for downstream tasks requiring human-like decision-making capacity and output is often studied~\citep{Wang2023ASO}. 
For example, planning is a hallmark of human intelligence, and thus a core requirement of such agentic systems. 
Hence, there is a considerable volume of research investigating the planning and reasoning capacity of LLMs;~\citet{10.5555/3666122.3667815} define several important subtasks, including task decomposition, goal reformulation, and plan generation, among others. In this regard, hallucination remains a persistent shortcoming, e.g. inconsistency in following complex task prompts, invalid planned actions, and otherwise illogically-formed plans~\citep{huang2024understandingplanningllmagents}.

\par Regardless, the specific application of LLMs to retrieval-augmented generation (RAG)-based SQA is underexplored, and there remain practical challenges associated with this task.

\par Firstly, regarding the \textbf{unique structure and phrasing of scientometric questions} - by definition, scientometric analysis typically concerns (groups of) specific academic entities, which influence the RAG database querying process. This separates them from standard, topic-based scientific queries~\citep{LI2021105093}. Hence, correct task decomposition and subtask planning is of particular significance.

\par In relation, \textbf{database querying can be complex and multi-faceted} when accounting for the added expressivity of scientometric questions; search query generation must encompass the various filtering and faceting options supported by the database querying language, which may be custom, i.e. non-standard, depending on the specific database used.
For example, the question may require an aggregation of entities, e.g. retrieving the top authors associated with publications of a particular entity. 
This further stresses the importance of effective plan generation and adherence to complex, highly-specific task prompts. 

\par Consequently, \textbf{solely relying on an LLM's parametric general knowledge}, e.g. using a direct generation setup, \textbf{is insufficient} for the nuances of such questions; the LLM may generate answers based on inaccurate and/or outdated data, or hallucinate fictitious references, which necessitates an efficient RAG interface~\citep{10.1145/3571730, NEURIPS2020_6b493230}. 

\par Given these interesting challenges, we implement \textbf{\textsc{AnalyticsGPT}}, an intuitive and reliable LLM workflow for SQA. The system is motivated by the following hypotheses: (1) LLMs perform better when given a high-level outline of the task (influenced by the chain-of-thought paradigm)~\citep{10.5555/3600270.3602070}, (2) LLMs perform better when handling a specialized task at a given time~\citep{zhang2024examinationeffectivenessdivideandconquerprompting}, and (3) LLMs are not good at generating directly-executable plans~\citep{10.5555/3666122.3667815}. Based on these hypotheses, we define five key modules with a sequential order of execution, illustrated in Figure~\ref{fig:sys_diag}.

\par As a RAG database, we utilize a proprietary research analytics platform; various application programming interface (API) endpoints serve as the entity lookup and database querying tools with which we resolve entity names to their corresponding identifiers (IDs) and query for data, respectively. 

\par Evaluation by subject matter experts (SMEs) and LLMs-as-judges reveals that \textsc{AnalyticsGPT} consistently outperforms a naive RAG-based LLM baseline in terms of robustness, content coverage, and response claim validity, without sacrificing speed, on an evaluation set consisting of user and synthetic questions. To the best of our knowledge, our method is the first to tackle the SQA task using LLMs.

\section{User Needs}\label{sec:user_needs}
\par The task concerns two key interconnected products within our business.
Consider Product A, an abstract and citation database, and Product B, a research analytics platform that primarily relies on the abstract and citation data from Product A; there is a significant overlap in the customer base for both products. Currently, customers using Product A who have analytics questions are directed via hyperlinks to Product B. However, Product B has a vastly different user experience compared to Product A and requires a separate login, which Product A does not enforce. This disjointed user experience often leads to new users abandoning their queries before finding an answer.
\par Our objective is to enhance the user experience by enabling users to answer simple metric questions without needing to switch products. The ability to perform adjacent tasks within a single product is common practice and leads to a more efficient workflow. In our case, a user may be examining a set of results and want to know which researchers have published the most in a certain research area. Instead of directing them to Product B, users can type their questions directly into Product A, and the relevant metrics will be retrieved from Product B. The answer will then be displayed within Product A, providing a seamless and intuitive solution. This approach enables natural language interaction with the data, thereby reducing the barrier to entry.

\section{Methodology}\label{sec:methodology}
We propose an approach centered around a fixed workflow, with each successive module executed in sequence. 
Figure~\ref{fig:sys_diag} provides a high-level overview of the system modules. 
The system implementation utilizes the LangChain framework.
Note that while we also provide details specific to our concrete implementation where relevant, the proposed general workflow can be instantiated with different implementation details as well.

\subsection{User Input}
In this subsection, we attempt to formalize the expressivity of the system, i.e. specifying the exact nature of user questions it is designed to handle.

\subsubsection{Supported Entities}\label{sec:supported_entities}
The system supports direct referencing of a variety of academic entity types, e.g. Authors, Institutions, Journals, Topics, Subject Areas, etc. Importantly, articles are not included, since they serve as the fundamental data unit of the RAG database; metadata from articles is aggregated to compute the requested data. Thus, they are implicitly used and are not directly referenced. 
Furthermore, the system is intended to be embedded within the interface of our abstract and citation database, which can already be used to look up articles by title.

\subsubsection{Question Forms}\label{sec:question_forms}
\par We adapt (a subset of) the definitions provided by~\citet{banerjee2023dblpquadquestionansweringdataset}, with some alterations. Not all questions of these forms are necessarily supported by the system - certain implementation constraints must be accounted for - but all questions supported by the system fall under these categories. 
\par \textbf{Fact-based.} Answerable with a single fact, requiring no data aggregation, e.g. \textit{``Mention the co-authors of Chang Yun Park.''} Additional filter parameters allow multiple-fact questions, e.g. a condition that the papers used must be from the past 3 years.
\par \textbf{Single-intent.} A single user intention over a single subject. The intention can also concern an entity, e.g. \textit{``Who are the most cited authors in the field of Neuroscience at the University of Oxford?''}
\par \textbf{Union.} Single-intent questions over multiple separate entities, e.g. \textit{``Marco D. Santambrogio and Durelli, G. have which primary affiliations?''}
\par \textbf{Multiple-intent.} Multiple user intentions over a single subject, e.g. \textit{``Which publications did Qiubin Gao author and in which year?''}
\par \textbf{Comparative \& superlative.} The former type compares values between multiple entities; note that the point of comparison can also be an entity, e.g. \textit{``How does the University of Tokyo's research output in Physics compare with that of the University of Cambridge?''} The latter type inquires about maximum values between entities, e.g. \textit{``Report the most frequent co-author of M. Schreuder and how many papers do they have together?''}
\par Note that although we use the term ``question,'' we do not differentiate between actual questions and their imperative statement forms. In this context, a user ``intent'' is any singular data aggregation request. Aside from the base question, additional filters are supported, including: publication year, metrics to retrieve (e.g. citation count, field-weighted citation impact), and faceted searches with respect to other entities. This scope was defined based on prior analysis of attempted user questions during the conceptual phase.

\subsection{RAG Interface and Tools}\label{sec:rag}
We programmatically access our proprietary research analytics platform through various API endpoints; to the LLM, these are represented as database querying tools. While~\citet{patil2024gorilla} have demonstrated that fine-tuned LLMs can handle tool calling with massive APIs, we maintain a high-level of abstraction by encapsulating numerous low-level endpoints within just a few tool implementations. In doing so, we shift the complexity of the task by trivializing the correct tool selection, and focusing on correct complex parameter generation for said tools instead. To elaborate, the following tools are defined and exposed to the system:
\par \textbf{Entity name $\rightarrow$ ID resolution.} Resolves entity names to their best-matching IDs, using a proprietary GraphQL layer aggregating various vector-based search endpoints.
\par \textbf{Article search.} Query for articles from the platform, optionally with filters. As introduced in Section~\ref{sec:question_forms}, a number of (optional) filtering parameters can be specified.
\par \textbf{Faceted article search.} Aggregates the top entities associated with a set of filtered articles, optionally retrieving metrics of said entities.
\par While an essential component of the HLPM, the entity resolution tool is used independently of any LLM calls. Thus, only two tools are exposed for planning purposes, both with a broad range of parameter values to specify.

\subsection{High-Level Planning Module}\label{sec:HLPM}
The HLPM serves as the entry point to the pipeline, and consists of two sub-components:
\par (1) An LLM call, in which the LLM is prompted to: (1.1) Perform NER on the user query, tagging supported entity types, and (1.2) decompose the user query into an abstract step-by-step outline; the model is instructed not to include specific tools, parameters, or otherwise low-level implementation details in this outline.
\par (2) Based on the LLM NER, resolve the tagged entity names to their corresponding IDs, for use in further data retrieval steps.
\par As described in Section~\ref{sec:rag}, our entity resolution implementation leverages a proprietary GraphQL layer. However, in principle, an alternative entity resolution method can be easily substituted in a different concrete implementation of the workflow we propose. The step-by-step outline and resolved entity IDs are provided as context to the DPM.

\subsection{Detailed Planning Module}\label{sec:DPM}
The primary task of the DPM is to elaborate upon the outline drafted by the HLPM, and consists of a single LLM call, with instructions to output a detailed, low-level plan consisting of a list of data retrieval steps necessary to address the query. 
\par Per step, the following details are specified: (1) The name of the tool to be called, (2) the subtask for which the tool is called, (3) dependencies on previous steps in the plan (if any), and (4) the parameter values with which the tool should be called.
\par To facilitate accurate specification of parameter values, the prompt includes method contracts and few-shot examples for the tools. Note that we do not instruct the model to directly generate the search query, as ad-hoc tests revealed occasional syntax errors, negatively impacting robustness. The DPM is critical to a high-quality final response, as it dictates both the extent and correctness of data points that can be later referenced by the WM. Hence, these design choices are taken to ensure consistent and expected behavior across runs.

\subsection{Action Module}\label{sec:aa}
Based on the detailed plan, the AM executes the tool calls using the stated parameters and collects the data per step. We perform situational LLM tool calling - for a given step, there are two possible modes of execution: (1) If the step is \textit{independent}, the requested tool is directly invoked with the given parameters, and (2) if the step is \textit{dependent}, an LLM tool call is first generated, as certain parameter values must be inferred from the results of previous data retrieval steps.
\par All independent steps are executed concurrently; doing so provides the most substantial time savings in the case of union questions, which are often embarrassingly parallel by definition. At each step, prior to tool invocation, the required database search query is assembled from the given parameters using a rule-based approach. This facilitates the formulation of syntactically-correct queries, as we can handle common mistakes by the LLM, and mitigates the risk of syntax error-induced crashes.

\subsection{Writing Module}\label{sec:wa}
The WM consists of a single LLM call, instructed to compose a final summarized response for the user, containing high-level analysis and insights based on the retrieved data. Due to the importance of providing factually-correct statements supported by references to the retrieved data, the model is prompted to not fill in data gaps with internal knowledge. This further stresses the aforementioned criticality of correct data retrieval in Section~\ref{sec:DPM}. The prompt emphasizes the use of in-line referencing and coherent structuring through the use of markdown tables, headings, and lists.

\subsection{Visualization Module}\label{sec:va}
The VM acts as an augmentation component after the WM. It utilizes a set of LLM calls and dynamic code execution. The purpose of the initial LLM call is deciding whether the final summarized response from the WM would be better described by additional plotting images, leading to an enhanced user experience~\citep{borkin2013makes, saket2016beyond}. In that case, the LLM processes the user's query with the final response as context to create appropriate visualization. LLMs have shown the ability of code comprehension and generation~\citep{jiang2024surveylargelanguagemodels}. The model instructions are to generate the code-structure for the plotting and then execute an internal Python routine to dynamically run the plot generation process and feed any errors that may occur back to the LLM for bug-fixing, until it can stream back the plot image(s). Plots are fitted in the context of the final response to the user and the data are visualized accordingly.

\section{System Evaluation}
We investigate system performance on the task of SQA. As a baseline for comparison, we also implement a naive RAG-based LLM approach that first performs tool selection based on the input user question, then constructs a final response based on the retrieved data; module-wise, the baseline shares (parts of) the AM and WM with \textsc{AnalyticsGPT}, but does not include the preparatory planning of the HLPM and DPM. Also, the simplified AM variant used in the baseline does \textit{not} feature the rule-based approach for manual database query building used by \textsc{AnalyticsGPT} (discussed in Section~\ref{sec:aa}). Thus, the baseline represents the simplest end-to-end RAG pipeline possible for handling SQA with respect to our product ecosystem - (multiple) tool calling followed by summarization.
Regarding models, we use GPT-4o for the DPM, and GPT-4o Mini for all other modules.

\subsection{Human Annotation}
For evaluation, we ask two SMEs to grade the quality of the generated response using a five-point rubric with multiple sub-indicators (seen in Appendix~\ref{apx:eval_crit}). The SMEs are recruited from a team of specialists, are experienced in the qualitative evaluation of (other) LLM solution concepts within our institution, and are familiar with the product ecosystem (discussed in Section \ref{sec:user_needs}) surrounding our system. 
We also manually inspect the input data to the WM for each sample to check for critical errors within the data retrieval phase.

\subsection{LLM Annotation}
LLMs have been successfully employed as evaluators across varied NLP tasks~\citep{gu2025surveyllmasajudge}; we further examine the efficacy of LLMs-as-judges given the complex rubric at hand. Rather than the judgment of individual LLMs, we implement an LLM ``jury,'' i.e. a panel of LLM evaluators, which has been shown to mitigate intra-model bias and increase agreement with human annotators~\citep{verga2024replacingjudgesjuriesevaluating}. As a voting function to pool individual model scores, we utilize \textit{confidence-weighted pooling}: we prompt the judges to provide a confidence (0-1) for their given score, and the score with the highest average confidence among all judges is taken. If there is no significant confidence difference between given scores, we fall back to simple majority voting. The jury consists of four models: GPT-4o Mini, GPT-4.1 Mini, Claude 3.5 Haiku, and Claude 3.5 Sonnet.

\subsection{Evaluation Dataset}\label{sec:eval_dataset}
We utilize a modest evaluation set of 84 scientometric questions, sourced from rounds of user testing (34 questions), as well as DBLP-QuAD by~\citet{banerjee2023dblpquadquestionansweringdataset}, a question answering dataset on the basis of bibliographic information from the DBLP scholarly knowledge graph. 
\begin{figure}[h]
    \centering
    \includegraphics[width=\linewidth]{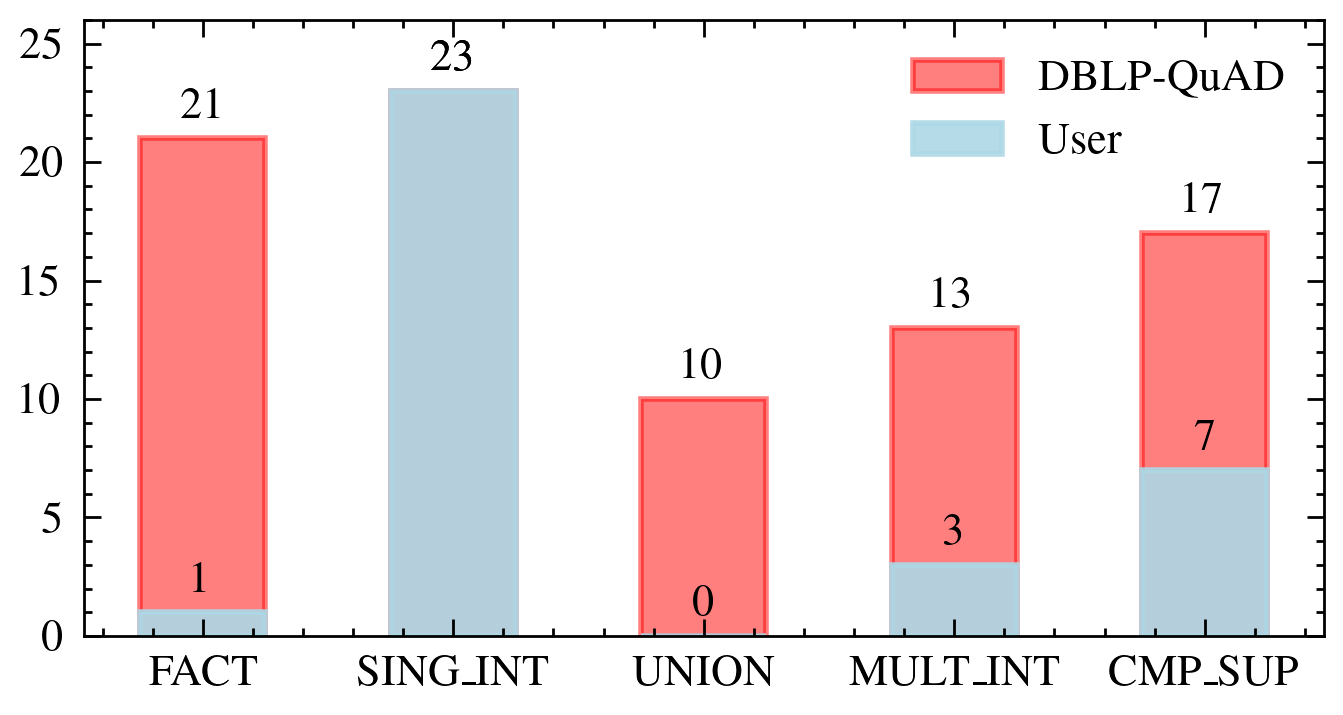}
    \caption{\label{fig:data_dist}Distribution of question forms by count in the evaluation set. Note that single-intent (SING\_INT) is a custom definition and not part of DBLP-QuAD. We overrepresent the fact-based category to pad the dataset with ample base cases, as users often tried to ask more complex questions.}
\end{figure}
\par In general, user questions tend to be more strategic and conversational in nature compared to DBLP-QuAD questions, which are synthetically generated based on a series of human-written templates. For DBLP-QuAD, we randomly sample 50 questions from the test set - 10 each from the following categories: single-fact, multi-fact, double-intent, union, and comparative \& superlative. Within these categories, certain question templates were incompatible with our system due to representation differences in the underlying databases, and were excluded from the sampling process.

\subsection{Results}\label{sec:results}
\begin{figure*}[t]
    \centering
    \includegraphics[width=\linewidth]{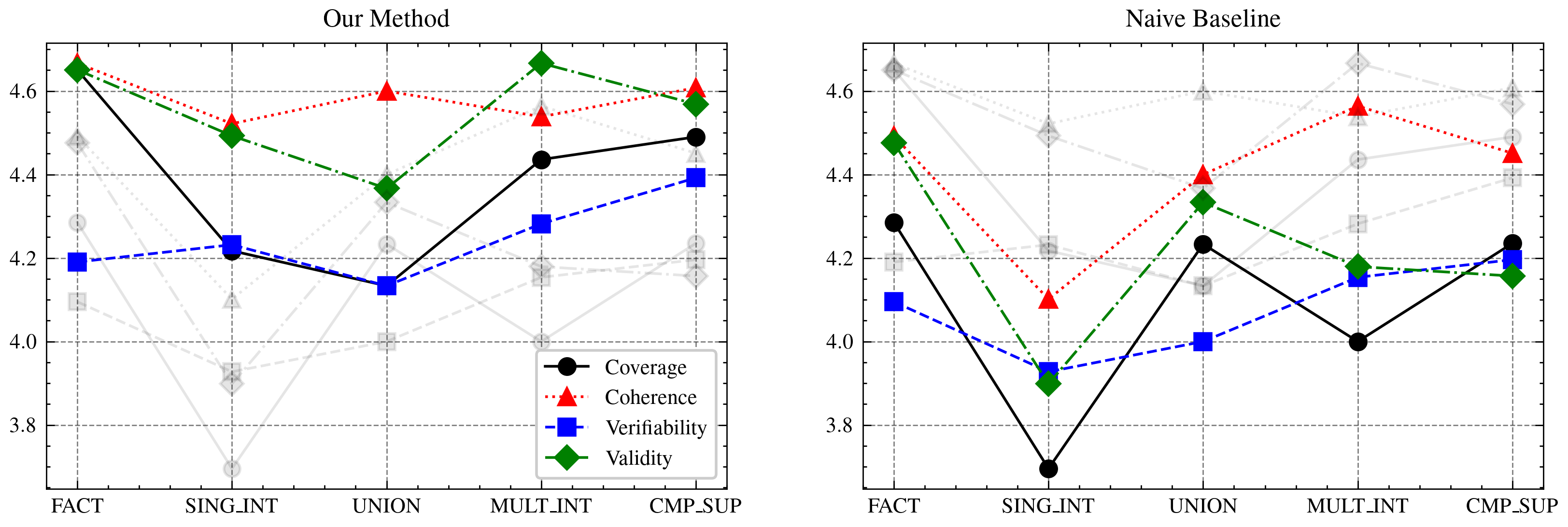}
    \caption{\label{fig:score_by_form}Criteria scores separated by question form (as described in Section~\ref{sec:question_forms}), per method. Each graph also indicates the trace of the other, for ease of comparison.}
\end{figure*}
Average evaluation scores for both methods are illustrated in Table~\ref{tab:results}, with separation by question form in Figure~\ref{fig:score_by_form}. While the baseline achieves respectable scores, results indicate that \textsc{AnalyticsGPT} significantly outperforms the baseline in Coverage and Validity, and with greater consistency. An output example can be seen in Appendix~\ref{apx:out_example}.

\begin{table}[h]
\centering
\resizebox{\linewidth}{!}{%
\begin{tabular}{@{}lll@{}}
\toprule
\multicolumn{1}{l|}{}                          & \multicolumn{1}{l|}{Naive Baseline}                & \textsc{AnalyticsGPT}                                                        \\ \midrule
\multicolumn{1}{l|}{\textbf{\# Resp. Tokens}}  & \multicolumn{1}{l|}{$\text{624}_{\text{± 258}}$}   & $\text{681}_{\text{± 322}}$                                         \\ \midrule
\multicolumn{1}{l|}{\textbf{API Time (s)}}         & \multicolumn{1}{l|}{$\text{14.2}_{\text{± 6.1}}$}  & $\text{20.9}_{\text{± 12.3}}$ \\ \midrule
\multicolumn{1}{l|}{\makecell[l]{\textbf{Critical} \textbf{Errors}}} & \multicolumn{1}{l|}{5/84}                          & 1/84                                                                \\ \midrule
\multicolumn{3}{c}{\textbf{SME \& LLM Evaluation}}                                                                                                                        \\ \midrule
\multicolumn{1}{l|}{\textit{Coverage}}         & \multicolumn{1}{l|}{$\text{4.06}_{\text{± 1.13}}$} & $\text{4.40}_{\text{± 0.95}}$ $\textcolor{green}{\blacktriangle}$   \\ \midrule
\multicolumn{1}{l|}{\textit{Coherence}}        & \multicolumn{1}{l|}{$\text{4.38}_{\text{± 1.00}}$} & $\text{4.59}_{\text{± 0.66}}$ $\textcolor{lightgray}{\blacktriangle}$   \\ \midrule
\multicolumn{1}{l|}{\textit{Verifiability}} & \multicolumn{1}{l|}{$\text{4.07}_{\text{± 1.01}}$} & $\text{4.25}_{\text{± 0.70}}$ $\textcolor{lightgray}{\blacktriangle}$ \\ \midrule
\multicolumn{1}{l|}{\textit{Validity}}         & \multicolumn{1}{l|}{$\text{4.19}_{\text{± 1.15}}$} & $\text{4.56}_{\text{± 0.75}}$ $\textcolor{green}{\blacktriangle}$   \\ \midrule
\multicolumn{1}{l|}{\textit{Avg.}}             & \multicolumn{1}{l|}{4.17}                          & 4.45                                                                \\ \bottomrule
\end{tabular}%
}
\caption{\label{tab:results}Average response length in tokens, time taken for all API calls in seconds, number of critical data retrieval errors, and criteria scores for both methods. Green markers indicate statistically significant increases at $\alpha = 0.05$. (See Appendix~\ref{apx:sig} for the \textit{p}-values.)}
\end{table}

\par Generally, the in-depth planning and guided database querying of \textsc{AnalyticsGPT} proves advantageous in this regard, consistently providing complete and relevant data. The ability to handle nested requests can also prompt higher Coverage for non-nested input questions, as dependent steps can retrieve relevant supporting data beyond the exact wording of the question (e.g. additional recent publications or citation counts). These advantages are reflected by the higher Coverage scores across multiple question forms, notably in the single-intent category, which consists entirely of user questions with more varied requests and less-concrete wording.
\subsection{Failure Modes}\label{sec:failure_modes}
\par Concerning retrieval errors, manual analysis uncovered critical data omission\footnote[1]{Cases in which no data was retrieved at all, while manual checks using correct queries did return usable data points.} in five out of 84 questions for the baseline (four of which were user questions); these were caused by parameterization errors during database query generation, e.g. conflating Topics and Subject Area identifiers, among other parameter misuse. Moreover, given the natural verbosity of LLMs when insufficient context is provided, validating claims based on the data becomes more challenging~\citep{saito2023verbositybiaspreferencelabeling, briakou2024implicationsverbosellmoutputs}; attempted explanatory statements can also cause the LLM to deviate from structuring and referencing guidelines. While \textsc{AnalyticsGPT} mitigates these issues on our modest evaluation set (thus achieving high Validity), more rigorous testing is required to determine their persistence.
\par Both systems occasionally struggled to capture important nuances in user questions. Notably, descriptive temporal modifiers, e.g. ``emerging,'' ``cutting-edge,'' that could influence the year range of search queries, were often lost in translation. Hallucinated references and in-text hyperlinks, e.g. fictional DOI links for article titles (even in the presence of correctly-retrieved article IDs), negatively influenced the Verifiability of both methods.

\section{Business Impact}\label{sec:business}
\par Streamlining the research analytics process through user-centric design and a conversational interface not only mitigates the steep learning curve typically associated with traditional analytics platforms but also enables research leaders to rapidly access high-level insights. \textsc{AnalyticsGPT} enables stakeholders to make informed decisions regarding resource allocation, strategic investments, and research collaborations. 
\par In pilot user studies, the system has effectively highlighted the contribution of specific research facilities to overall institutional performance, as well as delineated the impact of individual research outputs relative to institutional averages. The automation of data synthesis has been observed to significantly decrease the time-to-insight. Looking ahead, \textsc{AnalyticsGPT} presents opportunities for revenue growth, particularly through upselling within our broader product ecosystem and targeting Tier 2 and Tier 3 institutions that may lack extensive analytics resources. 

\section{Conclusion and Future Work}
In this paper, we describe our implementation of an LLM-driven, RAG-based workflow for SQA. We leverage a proprietary research analytics platform, and focus on effective task decomposition and database query construction to guide the data retrieval process. We also concentrate on presenting findings with well-structured responses, verifiable insights and analyses, and metrics visualization. In so doing, we develop a system that effectively provides access to high-level scientometric insights through a natural language conversational interface embedded within our proprietary abstract and citation database. Results from combined SME and LLM evaluation suggest consistent improvement over a naive RAG-based pipeline in terms of Coverage and Validity, on both user and synthetic questions. Importantly, initial user feedback indicated a marked decrease in time-to-insight, adding value to the user journey.
\par Future developments include the development of an automated framework for effective scalable evaluation, and investigating the performance of large reasoning models for detailed reflective plan generation. Areas of improvement include hallucinated references and response trustworthiness, translation of nuanced user requests, and general expressivity of the system (and supporting RAG interface) to handle increasingly complex user questions.

\section*{Limitations}\label{sec:limitations}
\par The defined evaluation criteria is fine-grained and tailored towards SME expertise and manual checks; this proved challenging for the LLMs-as-judges, even in a jury setting, which scored responses leniently in comparison, causing low agreement (see Appendix~\ref{apx:iaa}). A reliable and scalable automated evaluation framework would strengthen the conclusions drawn. Furthermore, the comparison would benefit from a use-case level evaluation focusing on possible failure modes during user trials; as mentioned in Section~\ref{sec:business}, this would better highlight the exact extent to which the system supports their high-level decision-making.
\par Finally, a more detailed ablation study investigating the individual impact of modules (e.g. DPM and HLPM) could provide better insight on the performance gains achieved.

\section*{Ethical Considerations}
\par The authors have full rights to the use of any proprietary tools discussed throughout this paper, e.g. the mentioned research analytics platform. In relation, given that the research was carried out using proprietary platforms, while we provide skeleton code and prompts, a fully-functioning prototype (i.e. with access to our APIs) cannot be made open-source.
\par The evaluation dataset used contains user questions acquired from closed testing, i.e. a restricted user-base with early access to the system. Personally-identifiable information (if any) was stripped from the questions.
\par The human annotators were fairly compensated for their expertise. Methods were presented anonymously, i.e. ``Method A,'' ``Method B,'' in identical format, to mitigate bias.

\bibliography{custom}

@article{Wang2023ASO,
      title={A Survey on Large Language Model based Autonomous Agents},
      author={Lei Wang and Chengbang Ma and Xueyang Feng and Zeyu Zhang and Hao-ran Yang and Jingsen Zhang and Zhi-Yang Chen and Jiakai Tang and Xu Chen and Yankai Lin and Wayne Xin Zhao and Zhewei Wei and Ji-rong Wen},
      journal={Frontiers Comput. Sci.},
      year={2023},
      volume={18},
      pages={186345},
      url={https://api.semanticscholar.org/CorpusID:261064713}
}

@misc{jiang2024surveylargelanguagemodels,
    title={A Survey on Large Language Models for Code Generation}, 
    author={Juyong Jiang and Fan Wang and Jiasi Shen and Sungju Kim and Sunghun Kim},
    year={2024},
    eprint={2406.00515},
    archivePrefix={arXiv},
    primaryClass={cs.CL},
    url={https://arxiv.org/abs/2406.00515}, 
}

@inproceedings{10.5555/3600270.3602070,
author = {Wei, Jason and Wang, Xuezhi and Schuurmans, Dale and Bosma, Maarten and Ichter, Brian and Xia, Fei and Chi, Ed H. and Le, Quoc V. and Zhou, Denny},
title = {Chain-of-thought prompting elicits reasoning in large language models},
year = {2022},
isbn = {9781713871088},
publisher = {Curran Associates Inc.},
address = {Red Hook, NY, USA},
booktitle = {Proceedings of the 36th International Conference on Neural Information Processing Systems},
articleno = {1800},
numpages = {14},
location = {New Orleans, LA, USA},
series = {NIPS '22}
}

@misc{zhang2024examinationeffectivenessdivideandconquerprompting,
      title={An Examination on the Effectiveness of Divide-and-Conquer Prompting in Large Language Models}, 
      author={Yizhou Zhang and Lun Du and Defu Cao and Qiang Fu and Yan Liu},
      year={2024},
      eprint={2402.05359},
      archivePrefix={arXiv},
      primaryClass={cs.AI},
      url={https://arxiv.org/abs/2402.05359}, 
}

@article{altbach2012,
author = {Altbach, Philip},
year = {2012},
month = {01},
pages = {26-31},
title = {The Globalization of College and University Rankings},
volume = {44},
journal = {Change: The Magazine of Higher Learning},
doi = {10.1080/00091383.2012.636001}
}

@book{hazelkorn2015,
author = {Hazelkorn, Ellen},
year = {2015},
month = {03},
pages = {},
title = {Rankings and the Reshaping of Higher Education. The Battle for World-Class Excellence},
isbn = {9781349318216},
doi = {10.1057/9780230306394}
}

@inproceedings{10.5555/3666122.3667815,
author = {Valmeekam, Karthik and Marquez, Matthew and Olmo, Alberto and Sreedharan, Sarath and Kambhampati, Subbarao},
title = {PlanBench: an extensible benchmark for evaluating large language models on planning and reasoning about change},
year = {2023},
publisher = {Curran Associates Inc.},
address = {Red Hook, NY, USA},
booktitle = {Proceedings of the 37th International Conference on Neural Information Processing Systems},
articleno = {1693},
numpages = {13},
location = {New Orleans, LA, USA},
series = {NIPS '23}
}

@misc{huang2024understandingplanningllmagents,
      title={Understanding the planning of LLM agents: A survey}, 
      author={Xu Huang and Weiwen Liu and Xiaolong Chen and Xingmei Wang and Hao Wang and Defu Lian and Yasheng Wang and Ruiming Tang and Enhong Chen},
      year={2024},
      eprint={2402.02716},
      archivePrefix={arXiv},
      primaryClass={cs.AI},
      url={https://arxiv.org/abs/2402.02716}, 
}

@article{gevorgyan2016,
author = {Gevorgyan, Ashot},
year = {2016},
month = {07},
pages = {29},
title = {Advanced Scientometric Databases as an Objective Criterion of Scientific Activities},
volume = {1},
journal = {wisdom},
doi = {10.24234/wisdom.v1i6.60}
}

@inproceedings{NEURIPS2020_6b493230,
	author = {Lewis, Patrick and Perez, Ethan and Piktus, Aleksandra and Petroni, Fabio and Karpukhin, Vladimir and Goyal, Naman and K\"{u}ttler, Heinrich and Lewis, Mike and Yih, Wen-tau and Rockt\"{a}schel, Tim and Riedel, Sebastian and Kiela, Douwe},
	booktitle = {Advances in Neural Information Processing Systems},
	editor = {H. Larochelle and M. Ranzato and R. Hadsell and M.F. Balcan and H. Lin},
	pages = {9459--9474},
	publisher = {Curran Associates, Inc.},
	title = {Retrieval-Augmented Generation for Knowledge-Intensive NLP Tasks},
	url = {https://proceedings.neurips.cc/paper_files/paper/2020/file/6b493230205f780e1bc26945df7481e5-Paper.pdf},
	volume = {33},
	year = {2020}}

@misc{banerjee2023dblpquadquestionansweringdataset,
      title={DBLP-QuAD: A Question Answering Dataset over the DBLP Scholarly Knowledge Graph}, 
      author={Debayan Banerjee and Sushil Awale and Ricardo Usbeck and Chris Biemann},
      year={2023},
      eprint={2303.13351},
      archivePrefix={arXiv},
      primaryClass={cs.DL},
      url={https://arxiv.org/abs/2303.13351}, 
}

@article{10.1145/3571730,
author = {Ji, Ziwei and Lee, Nayeon and Frieske, Rita and Yu, Tiezheng and Su, Dan and Xu, Yan and Ishii, Etsuko and Bang, Ye Jin and Madotto, Andrea and Fung, Pascale},
title = {Survey of Hallucination in Natural Language Generation},
year = {2023},
issue_date = {December 2023},
publisher = {Association for Computing Machinery},
address = {New York, NY, USA},
volume = {55},
number = {12},
issn = {0360-0300},
url = {https://doi.org/10.1145/3571730},
doi = {10.1145/3571730},
journal = {ACM Comput. Surv.},
month = mar,
articleno = {248},
numpages = {38}
}

@article{LI2021105093,
	author = {Jie Li and Floris Goerlandt and Genserik Reniers},
	doi = {https://doi.org/10.1016/j.ssci.2020.105093},
	issn = {0925-7535},
	journal = {Safety Science},
	pages = {105093},
	title = {An overview of scientometric mapping for the safety science community: Methods, tools, and framework},
	url = {https://www.sciencedirect.com/science/article/pii/S0925753520304902},
	volume = {134},
	year = {2021}}

@inproceedings{
patil2024gorilla,
title={Gorilla: Large Language Model Connected with Massive {API}s},
author={Shishir G Patil and Tianjun Zhang and Xin Wang and Joseph E. Gonzalez},
booktitle={The Thirty-eighth Annual Conference on Neural Information Processing Systems},
year={2024},
url={https://openreview.net/forum?id=tBRNC6YemY}
}

@article{borkin2013makes,
  title={What makes a visualization memorable?},
  author={Borkin, Michelle A and Vo, Azalea A and Bylinskii, Zoya and Isola, Phillip and Sunkavalli, Shashank and Oliva, Aude and Pfister, Hanspeter},
  journal={IEEE transactions on visualization and computer graphics},
  volume={19},
  number={12},
  pages={2306--2315},
  year={2013},
  publisher={IEEE}
}

@inproceedings{saket2016beyond,
  title={Beyond usability and performance: A review of user experience-focused evaluations in visualization},
  author={Saket, Bahador and Endert, Alex and Stasko, John},
  booktitle={Proceedings of the sixth workshop on beyond time and errors on novel evaluation methods for visualization},
  pages={133--142},
  year={2016}
}

@misc{briakou2024implicationsverbosellmoutputs,
      title={On the Implications of Verbose LLM Outputs: A Case Study in Translation Evaluation}, 
      author={Eleftheria Briakou and Zhongtao Liu and Colin Cherry and Markus Freitag},
      year={2024},
      eprint={2410.00863},
      archivePrefix={arXiv},
      primaryClass={cs.CL},
      url={https://arxiv.org/abs/2410.00863}, 
}

@misc{saito2023verbositybiaspreferencelabeling,
      title={Verbosity Bias in Preference Labeling by Large Language Models}, 
      author={Keita Saito and Akifumi Wachi and Koki Wataoka and Youhei Akimoto},
      year={2023},
      eprint={2310.10076},
      archivePrefix={arXiv},
      primaryClass={cs.CL},
      url={https://arxiv.org/abs/2310.10076}, 
}

@misc{gu2025surveyllmasajudge,
      title={A Survey on LLM-as-a-Judge}, 
      author={Jiawei Gu and Xuhui Jiang and Zhichao Shi and Hexiang Tan and Xuehao Zhai and Chengjin Xu and Wei Li and Yinghan Shen and Shengjie Ma and Honghao Liu and Saizhuo Wang and Kun Zhang and Yuanzhuo Wang and Wen Gao and Lionel Ni and Jian Guo},
      year={2025},
      eprint={2411.15594},
      archivePrefix={arXiv},
      primaryClass={cs.CL},
      url={https://arxiv.org/abs/2411.15594}, 
}

@misc{verga2024replacingjudgesjuriesevaluating,
      title={Replacing Judges with Juries: Evaluating LLM Generations with a Panel of Diverse Models}, 
      author={Pat Verga and Sebastian Hofstatter and Sophia Althammer and Yixuan Su and Aleksandra Piktus and Arkady Arkhangorodsky and Minjie Xu and Naomi White and Patrick Lewis},
      year={2024},
      eprint={2404.18796},
      archivePrefix={arXiv},
      primaryClass={cs.CL},
      url={https://arxiv.org/abs/2404.18796}, 
}

\newpage

\appendix

\section{Evaluation Criteria}\label{apx:eval_crit}
Table~\ref{tab:eval} shows the scoring rubrics used for system evaluation.


\section{Result Significance}\label{apx:sig}
Table~\ref{tab:sig} shows the unpaired two-tailed Mann-Whitney \textit{U}-test results (as the scores are non-normally distributed) for the criteria scores by method reported in Table~\ref{tab:results}.
\begin{table}[h]
\resizebox{\linewidth}{!}{%
\begin{tabular}{@{}l|llll@{}}
\toprule
        & Coverage & Coherence & Verifiability & Validity \\ \midrule
$U$       & 37821.5   & 34180.0    & 33674.5        & 36446.5   \\
\textit{p}-value & 4.189$e$-5   & 0.0817    & 0.1971        & 0.0009   \\ \bottomrule
\end{tabular}%
}
\caption{\textit{p}-values of reported scores. Values indicate a significant difference in Coverage and Validity at $\alpha = 0.05$.}
\label{tab:sig}
\end{table}

\section{Inter-Annotator Agreement}\label{apx:iaa}
Table~\ref{tab:iaa} shows the inter-annotator agreement between all annotators, separated by method and criteria. In general, while the human annotators exhibit fair to moderate agreement, there is heavy disagreement when factoring in the LLM jury, which tended to be much more lenient. This illustrates the limitations of using LLMs as judges, notably when evaluating nuanced criteria with in-depth requirements.
\begin{table}[h]
\resizebox{\linewidth}{!}{%
\begin{tabular}{@{}llllll@{}}
\toprule
                                                       & Method                & Coverage & Coherence & Verifiability & Validity \\ \midrule
\multicolumn{1}{l|}{\multirow{2}{*}{SME 1 - SME 2}}    & \textsc{AnalyticsGPT} & 0.215    & 0.209     & 0.342         & 0.141    \\ \cmidrule(l){2-6} 
\multicolumn{1}{l|}{}                                  & Naive Baseline        & 0.462    & 0.543     & 0.482         & 0.465    \\ \midrule
\multicolumn{1}{l|}{\multirow{2}{*}{SME 1 - LLM Jury}} & \textsc{AnalyticsGPT} & 0.18     & 0.0       & 0.201         & 0.0      \\ \cmidrule(l){2-6} 
\multicolumn{1}{l|}{}                                  & Naive Baseline        & 0.002    & -0.049    & 0.031         & -0.065   \\ \midrule
\multicolumn{1}{l|}{\multirow{2}{*}{SME 2 - LLM Jury}} & \textsc{AnalyticsGPT} & 0.076    & 0.0       & 0.084         & 0.0      \\ \cmidrule(l){2-6} 
\multicolumn{1}{l|}{}                                  & Naive Baseline        & 0.085    & 0.014     & 0.02          & 0.02     \\ \bottomrule
\end{tabular}
}
\caption{Inter-annotator agreement, shown pairwise (quadratic-weighted Cohen's kappa).}
\label{tab:iaa}
\end{table}

\section{Output Example}\label{apx:out_example}

\begin{table*}[h]
\resizebox{\textwidth}{!}{%
\begin{tabular}{@{}ll@{}}
\toprule
\textbf{Criterion} &
  \textbf{Scoring} \\ \midrule
\textbf{Coverage} &
  \begin{tabular}[c]{@{}l@{}}\textit{1}: Limited coverage; disregards (a) key aspect(s) of the user question.\\ \textit{2}: Partial coverage; lacking some discussion on aspects of the user question.\\ \textit{3}: Decent coverage; addresses all aspects of the user question to some degree, but some nuances are not (or incorrectly) accounted for.\\ \textit{4}: Complete coverage; addresses all aspects of the user question, effectively accounting for ambiguous requests within the question, \\ translating them into data requests compatible with the system.\\ \textit{5}: Above and beyond coverage; complete coverage of the user question, while providing additional relevant insights, suggestions, \\ and analysis on the retrieved data within scope.\end{tabular} \\ \midrule
\textbf{Coherence} &
  \begin{tabular}[c]{@{}l@{}}\textit{1}: Incoherent response; lacks logical structure and readability. Inappropriate or absent use of formatting.\\ \textit{2}: Limited coherence; somewhat logically structured and readable. Inappropriate use of formatting.\\ \textit{3}: Decent coherence; response flows logically and reads well. Formatting rules are applied to organize information, but there remain issues, \\ e.g. uninformative section headings or misuse of tables.\\ \textit{4}: Good coherence; response flows logically, reads well, and makes use of formatting rules to effectively present information where appropriate.\\ \textit{5}: Excellent coherence; response is optimally structured and reads well, presenting high-level insights, followed by factual data and analysis, \\ and concluding statements and references, respectively. Organization is spot-on, e.g. informative (sub-)sections, appropriate paragraph lengths, \\ effective use of tables and lists, etc.\end{tabular} \\ \midrule
\textbf{Verifiability} &
  \begin{tabular}[c]{@{}l@{}}\textit{1}: No supporting references are provided; claims are unsubstantiated.\\ \textit{2}: References are rarely provided; majority of claims are unsubstantiated.\\ \textit{3}: References are inconsistently provided; some of the claims remain unsubstantiated.\\ \textit{4}: References (both dedicated reference section and in-text hyperlinks) are generally provided where relevant; \\ majority of claims are supported by the retrieved data. \\ \textit{5}: References (both dedicated reference section and in-text hyperlinks) are always provided where relevant; \\ all claims are supported by the retrieved data, demonstrating high-level verification.\end{tabular} \\ \midrule
\textbf{Validity} &
  \begin{tabular}[c]{@{}l@{}}\textit{1}: Invalid; main claims and conclusions are based on complete misinterpretation or ignorance of the input data.\\ \textit{2}: Lacking validity; some claims and conclusions based on partial misinterpretation of the input data.\\ \textit{3}: Somewhat valid; main claims are connected to the input data, but some analytical and extrapolated statements are illogical.\\ \textit{4}: Generally valid; most claims made can be reasonably drawn from the input data. Most provided relevant data are present in the response, \\ but some potentially useful data are omitted.\\ \textit{5}: Fully valid; all claims made are correct interpretations based on the input data, and all provided relevant data are used appropriately in the response.\end{tabular} \\ \bottomrule
\end{tabular}%
}
\caption{\label{tab:eval}Scoring rubrics for SME evaluation.}
\end{table*}

Figure~\ref{fig:output_example} illustrates the markdown-formatted textual component of an example response from \textsc{AnalyticsGPT} for the question \textit{``What are the most relevant scholar contributions to SDGs which I should present in my `Sustainable Energy' day at school?''} SDG refers to the UN Sustainable Development Goals, a supported academic entity type for our system.
\begin{figure*}[ht]
\begin{lrbox}{\mybox}
    \begin{lstlisting}
    ### Summary of Relevant Scholarly Contributions to Sustainable Energy and SDGs
    
    For your 'Sustainable Energy' day at school, here are some of the most significant scholarly contributions related to Sustainable Development Goals (SDGs) in the context of sustainable energy. The insights are derived from recent publications and highlight key authors, institutions, and impactful articles.
    
    #### Key Contributions to Sustainable Development Goals (SDGs)
    
    1. **Top SDGs Related to Sustainable Energy**:
       - **SDG 7: Affordable and Clean Energy**
         - **Document Count**: 1,102
         - **Total Citations**: 9,681
         - **Average FWCI**: 0.63
       - **SDG 13: Climate Action**
         - **Document Count**: 664
         - **Total Citations**: 7,901
         - **Average FWCI**: 0.69
       - **SDG 9: Industry, Innovation, and Infrastructure**
         - **Document Count**: 395
         - **Total Citations**: 2,398
         - **Average FWCI**: 0.56
       - **SDG 8: Decent Work and Economic Growth**
         - **Document Count**: 360
         - **Total Citations**: 4,481
         - **Average FWCI**: 0.59
       - **SDG 12: Responsible Consumption and Production**
         - **Document Count**: 197
         - **Total Citations**: 3,344
         - **Average FWCI**: 0.47
    
       | SDG | Document Count | Total Citations | Average FWCI |
       |-----|----------------|------------------|--------------|
       | [SDG 7](SDG/SDG_v3_7) | 1,102 | 9,681 | 0.63 |
       | [SDG 13](SDG/SDG_v3_13) | 664 | 7,901 | 0.69 |
       | [SDG 9](SDG/SDG_v3_9) | 395 | 2,398 | 0.56 |
       | [SDG 8](SDG/SDG_v3_8) | 360 | 4,481 | 0.59 |
       | [SDG 12](SDG/SDG_v3_12) | 197 | 3,344 | 0.47 |
    
    #### Notable Authors and Their Contributions
    
    1. **Abdeen Mustafa Omer**
       - **Total Documents**: 152
       - **Total Citations**: 2,775
       - **Average FWCI**: 0.43
       - **Recent Works**:
         - [Energy use, environment and sustainable development](Paper/XXXXXXXXXXX) - 10 citations
         - [Development of sustainable energy research and applications](Paper/XXXXXXXXXXX) - 7 citations
    
    2. **Woodrow W. Clark**
       - **Total Documents**: 80
       - **Total Citations**: 585
       - **Average FWCI**: 0.38
       - **Recent Works**:
         - [A green hydrogen economy](Paper/XXXXXXXXXXX) - 107 citations
         - [Renewable energy and sustainable development](Paper/XXXXXXXXXXX) - 6 citations
    
    3. **Piotr Olczak**
       - **Total Documents**: 38
       - **Total Citations**: 490
       - **Average FWCI**: 1.08
       - **Recent Works**:
         - [Monofacial and bifacial micro pv installation as element of energy transition](Paper/XXXXXXXXXXX) - 40 citations
         - [The impact of economic, energy, and environmental factors on the development of the hydrogen economy](Paper/XXXXXXXXXXX) - 42 citations
    
    4. **Franco Fernando Yanine**
       - **Total Documents**: 21
       - **Total Citations**: 309
       - **Average FWCI**: 1.08
       - **Recent Works**:
         - [A literature review on an IoT-based intelligent smart energy management systems for PV power generation](Paper/XXXXXXXXXXX) - 20 citations
         - [Design of Smart Socket for Monitoring of IoT-Based Intelligent Smart Energy Management System](Paper/XXXXXXXXXXX) - 15 citations
    
    #### Conclusion
    
    These contributions highlight the ongoing research and advancements in sustainable energy, particularly in relation to the SDGs. Presenting these insights will not only inform your audience about the importance of sustainable energy but also inspire them to engage with these critical global challenges.
    
    ### References
    - **Topics**: [Sustainable Energy](Topic/XXXXX), [Renewable Energy](Topic/XXXXX), [Energy Management](Topic/XXXXX), [Energy Policy](Topic/XXXXX), [Energy Transition](Topic/XXXXX)
    - **SDGs**: [Affordable and Clean Energy](SDG/SDG_v3_7), [Climate Action](SDG/SDG_v3_13), [Industry, Innovation, and Infrastructure](SDG/SDG_v3_9), [Decent Work and Economic Growth](SDG/SDG_v3_8), [Responsible Consumption and Production](SDG/SDG_v3_12)
    \end{lstlisting}
\end{lrbox}
\centering
\scalebox{1}{\usebox{\mybox}}
\caption{\label{fig:output_example}Example textual response for \textit{``What are the most relevant scholar contributions to SDGs which I should present in my `Sustainable Energy' day at school?''} (Note that we have manually censored the IDs within in-text citations to papers; these were correctly cited by the model.)}
\end{figure*}

\end{document}